%% file: main.tex
\pdfoutput=1

\documentclass[journal]{IEEEtran}
\usepackage{cite}

\ifCLASSINFOpdf
  \usepackage[pdftex]{graphicx}
\else
\fi
\usepackage{amsmath}
  \usepackage[caption=false,font=footnotesize]{subfig}
\usepackage{url}

\input{macro.tex}

\begin{document}
%
\title{Fast Object Detection with Latticed Multi-Scale Feature Fusion}
%
%
%

\author{
    Yue~Shi$^*$,
    Bo~Jiang$^*$,
    Zhengping~Che, 
    and~Jian~Tang,~\IEEEmembership{Fellow,~IEEE}
    \thanks{
        $^*$The first two authors contributed equally to this work.
        Y. Shi, B. Jiang, Z. Che, and J. Tang are with AI Labs, Didi Chuxing, China.
        Z. Che is the corresponding author. Email: chezhengping@didiglobal.com
    }
}

%
%

\markboth{Fast Object Detection with Latticed Multi-Scale Feature Fusion}{}
%



\maketitle

\begin{abstract}
\input{abstract.tex}
\end{abstract}


%
\IEEEpeerreviewmaketitle

%
%
%
%



\section{Introduction}
\input{intro.tex}

\section{Related Work}
\input{related.tex}

\section{Methodology}
\input{method.tex}

\section{Experiments}
\label{sec:exp}
\input{exp.tex}

\section{Conclusion}
\input{conclusion.tex}

\ifCLASSOPTIONcaptionsoff
  \newpage
\fi


\bibliographystyle{IEEEtran}
\bibliography{reference}

\end{document}

%% file: macro.tex
\usepackage{booktabs}                                  
\usepackage{bbding,pifont}                             
\usepackage{multirow}                         		   
\usepackage{algorithm,algpseudocode}  				   
\usepackage{color,xcolor}							   

\newcommand{\ourmethod}{{Fluff}}
\newcommand{\ournet}{{\ourmethod}Net}

\newcommand{\eat}[1]{}                                 

\newcommand{\tinfer}{T_{\mathrm{infer}}}
\newcommand{\tpred}{T_{\mathrm{pred}}}
\newcommand{\tnms}{T_{\mathrm{nms}}}

%% file: abstract.tex
Scale variance is one of the crucial challenges in multi-scale object detection. Early approaches address this problem by exploiting the image and feature pyramid, which raises suboptimal results with computation burden and constrains from inherent network structures. Pioneering works also propose multi-scale (i.e., multi-level and multi-branch) feature fusions to remedy the issue and have achieved encouraging progress. However, existing fusions still have certain limitations such as feature scale inconsistency, ignorance of level-wise semantic transformation, and coarse granularity.
In this work, we present a novel module, the Fluff block, to alleviate drawbacks of current multi-scale fusion methods and facilitate multi-scale object detection. Specifically, {\ourmethod} leverages both multi-level and multi-branch schemes with dilated convolutions to have rapid, effective and finer-grained feature fusions. Furthermore, we integrate {\ourmethod} to SSD as {\ournet}, a powerful real-time single-stage detector for multi-scale object detection. Empirical results on MS COCO and PASCAL VOC have demonstrated that {\ournet} obtains remarkable efficiency with state-of-the-art accuracy. Additionally, we indicate the great generality of the {\ourmethod} block by showing how to embed it to other widely-used detectors as well. 

%% file: intro.tex
\input{intro-overview}

\input{intro-traditional-fusion-method.tex}

\input{intro-problems.tex}

\input{intro-architecture.tex}

\begin{figure}[t!]
  \centering
  \includegraphics[width=0.9\linewidth]{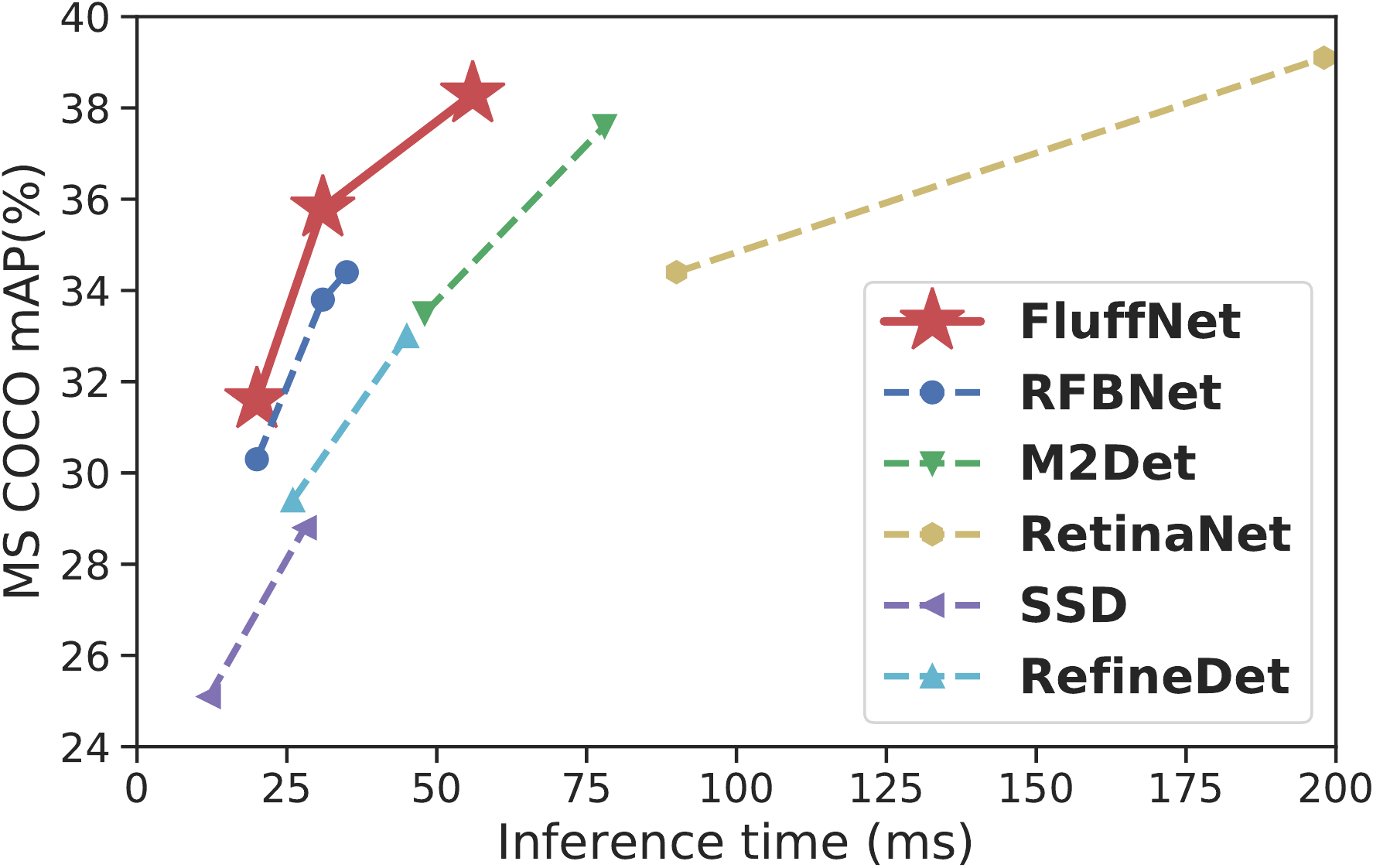}
  \caption{Speed (ms) vs. accuracy (mAP) on MS COCO \textit{test-dev}.}
  \label{fig:map-time}
\end{figure}

\input{intro-detector.tex}

\input{intro-contribution.tex}

%% file: intro-overview.tex
Object detection is one of the most significant tasks in computer vision. Recently, as an essential prerequisite for various downstream vision-based applications~\cite{He_2017_ICCV,liang2018cirl},
object detection has achieved great progress on prevailing benchmarks such as MS COCO~\cite{lin2014microsoft} and PASCAL VOC~\cite{everingham2007pascal} due to the rapid development of the deep convolution neural networks (DCNNs).
Particularly, there are two representative lines of approaches: two-stage methods such as the R-CNN series~\cite{Girshick_2015_ICCV,NIPS2015_5638}, and one-stage methods such as SSD~\cite{liu2016ssd} and YOLO~\cite{Redmon_2016_CVPR}.
Nevertheless, as a key challenge for object detection, scale variation across multi-scale instances stubbornly plays as a crucial issue in the above detectors, which encumbers detectors on extreme cases. 

%% file: intro-traditional-fusion-method.tex
To tackle the scale variation, pioneering works have raised multiple solutions. An instinctive one utilizes image pyramids~\cite{adelson1984pyramid} and explicitly exploits resized copies of images with diverse granularities on each scale. Related works~\cite{lowe2004distinctive,Singh_2018_CVPR} have clearly evidenced that DCNNs can be boosted by conducting this operation, while this scheme greatly increases memory usage and lacks computational efficiency.
In addition, detectors like SSD and FPN~\cite{Lin_2017_CVPR} achieve the scale invariability by feature pyramid. Feature maps extracted at different levels correspond to different scales for multi-scale predictions.
Though the feature pyramid reduces the computational complexity and can be easily integrated into DCNNs as end-to-end detectors, this method is still restrained by inherent structures as feature maps of each scale come solely from single-level layers which constrains the representational power of detectors.

Towards further enhancing object detections, multi-scale feature fusion is introduced, which roughly falls into two types: multi-level and multi-branch feature fusions. 
Detectors with multi-level feature fusions such as M2Det~\cite{zhao2019m2det} augment the features of shallow and deep layers to perform level-wise fusion.
Recently, EfficientDet~\cite{tan2020efficientdet} proposes a weighted bi-directional feature pyramid network to enhance the distinction of cross-scale feature and Libra R-CNN~\cite{pang2019libra} combines a balanced feature pyramid to strengthen the multi-level features.
Meanwhile, multi-branch feature fusions are equipped with parallel convolution branches. Particularly, TridentNet~\cite{li2019scale} uses three dilated convolution branches with weights sharing and RFBNet~\cite{liu2018receptive} builds on top of Inception~\cite{szegedy2015going} with varying sized dilated convolutions.

%% file: intro-problems.tex
Although both lines of effort for multi-scale feature fusion have achieved encouraging progress on multi-scale object detection, they still have certain drawbacks.
Multi-level feature fusion is impeded by inherent defects of the pyramidal design.
Specifically, the fused feature maps are not fine-grained enough since they are produced by size-specific layers and receptive fields that cannot be altered dramatically, leading to a confined granularity of scales for detection.
Thus, it is suboptimal to apply multi-level feature fusion to arbitrary objects as scales have wide and intensive distributions regarding categories and scenarios.
Differently, the multi-branch scheme effectively broadens the scale variations by adapting varying sized convolutions in parallel branches. However, it overlooks semantic transformation through vertical connections and neglects the shot to incorporate representations at different levels for objects with simple and complex appearances. Additionally, most multi-branch ones confront the problem of heavy computation when adopting larger kernel sizes for greater receptive fields with substantial inference cost. 
Relatively speaking, previous multi-scale fusion methods are \textit{coarse} as they simulate the multi-scale fusion in a sloppy way.
These detectors employ existing fusion methods on multiple crucial feature maps of different sizes once to fuse them at \textit{body level}, which annihilates the discriminative power of different feature maps for encoding the multi-scale instances.

%% file: intro-architecture.tex
In this paper, we propose a novel approach named \textbf{F}ast object detection with \textbf{L}atticed m\textbf{U}lti-scale \textbf{F}eature \textbf{F}usion (\textbf{\ourmethod}) to efficiently and effectively alleviate the above limitations. Concretely, {\ourmethod} leverages multi-level and multi-branch feature fusions through cascaded and parallel structures simultaneously in a latticed mode to incorporate the best of two worlds instead of applying them independently. 
As presented in Fig.~\ref{fig:block}, the {\ourmethod} block has multiple branches, each of which contains multiple levels of dilated convolutions~\cite{yu2015multi} of different rates. 
{\ourmethod} with varying sized receptive fields coherently energizes multi-scale feature fusion with finer granularity and flexibly lifts the fusion process from the network's structures.
Besides, multi-level design assists multi-branch fusion to maintain the semantic transformation through levels and strengthens the representational ability for objects with diverse appearances.
Meanwhile, dilated convolutions remarkably expedite the computation to achieve the fast inference. 
Last but not least, the {\ourmethod} blocks function on multiple key feature maps separately to perform \textit{head-level} fusion where each block only focuses on feature maps of a certain size rather than roughly fuse all maps together, which further advances {\ourmethod} for comprehensive multi-scale feature fusion. In brief, {\ourmethod} is a fast yet powerful block with \textit{fine-grained} multi-scale feature fusion. 

%% file: intro-detector.tex
In order to verify the validity and feasibility of the {\ourmethod} block, we embed it into standard SSD as a state-of-the-art end-to-end real-time detector named \textbf{{\ournet}}.
Extensive experiments conducted on MS COCO and PASCAL VOC have demonstrated its effectiveness. More importantly, our proposed block can be easily migrated to other widely-used detectors without losing generality. Our contributions can be summarized as follows:

%% file: intro-contribution.tex
\begin{itemize}
  \item We propose a novel module \textbf{{\ourmethod}} established with multi-level multi-branch fusions and dilated convolutions for the fast and fine-grained multi-scale feature fusion.
  \item We adapt {\ourmethod} to SSD to build \textbf{{\ournet}}, which obtains the state-of-the-art performance on MS COCO and PASCAL VOL, e.g., it inferences as fast as 32 FPS while achieving the mAP of 35.8\% on the MS COCO \textit{test-dev} set. Fig.~\ref{fig:map-time} depicts the speed-accuracy curves which demonstrates the superiority of {\ournet}.
  \item We present that {\ourmethod} can be easily integrated into other detectors as an essential and comprehensive module to boost the ability of multi-scale object detection.
\end{itemize}

%% file: related.tex
\paragraph{Object detectors}
DCNNs have become the protagonists on the stage of object detection, and current deep detectors are primarily divided into two categories: two-stage and one-stage ones. Two-stage detectors, like R-CNN series~\cite{Girshick_2014_CVPR,Girshick_2015_ICCV,NIPS2015_5638,He_2017_ICCV,li2017light} and TridentNet~\cite{li2019scale}, propose regions of interest first and feed extracted regional features to the following refinement layers. Despite the high accuracy, two-stage detectors always suffer from computational inefficiency. To accelerate the inference, one-stage detectors (e.g., SSD~\cite{liu2016ssd} and YOLO~\cite{Redmon_2016_CVPR,Redmon_2017_CVPR}) are presented to directly exploit the entire image in a feedforward manner without region proposals. Moreover, advanced one-stage detectors (e.g., RFBNet~\cite{liu2018receptive}, RefineDet~\cite{Zhang_2018_CVPR}, and M2Det~\cite{zhao2019m2det}) are also introduced to enhance the performance. Our work mainly focuses on one-stage detectors for the sake of efficiency.

\paragraph{Scale variation}
Scale variation has always been a key challenge in object detection. Plenty of works have been conducted to distance the effects of scale variation from detectors.
The image pyramid~\cite{adelson1984pyramid,lowe2004distinctive,dalal2005histograms,Singh_2018_CVPR} is an intuitive way to address scale variation by utilizing resized copies of input images, while it brings computation burdens.
Later, the feature pyramid~\cite{Lin_2017_CVPR} is employed on the single-scale image with improved efficiency.
However, it is essentially limited by the inherent structure of networks. 
Furthermore, works on multi-scale feature fusion including multi-level~\cite{2017arXiv170106659F,Lin_2017_CVPR,zhao2019m2det}
and multi-branch~\cite{szegedy2015going,szegedy2016rethinking,szegedy2017inception,chen2017rethinking,liu2018receptive,li2019scale}
feature fusions are invented and already show the promising progress on the task of multi-scale object detection.
Recently,
Libra R-CNN~\cite{pang2019libra} designs the balanced feature pyramid to strengthen the multi-level features using the same deeply integrated balanced semantic features.
HRNet~\cite{sun2019deep} and HigherHRNet~\cite{cheng2020higherhrnet} leverage multiple parallel high-to-low subnetworks with connections to conduct the multi-scale feature fusion.
EfficientDet~\cite{tan2020efficientdet} proposes a weighted bi-directional feature pyramid network and a compound scaling method.
Light-weighted and pyramid scale-equalizing convolutions~\cite{wang2020scale} are simultaneously developed to cater for inter-scale correlation.
Admittedly, the above fusion schemes carry with undeniable confinements. Multi-level fusions are constrained by the inherent structure of networks with feature scale inconsistency, and multi-branch approaches ignore the semantic transformation through layers.

\paragraph{Dilated convolution}
Dilated convolution~\cite{yu2015multi} effectively generates multi-scale receptive fields with different dilated rates on single feature maps and ensures the number of parameters to be constant simultaneously. It is widespread in semantic segmentation~\cite{chen2017deeplab,chen2017rethinking} and object detection~\cite{li2018detnet,liu2018receptive,li2019scale} since it can obtain broader context information while maintaining the high efficiency.

%% file: method.tex
\begin{figure}[t!]
  \centering
  \includegraphics[width=0.9\linewidth]{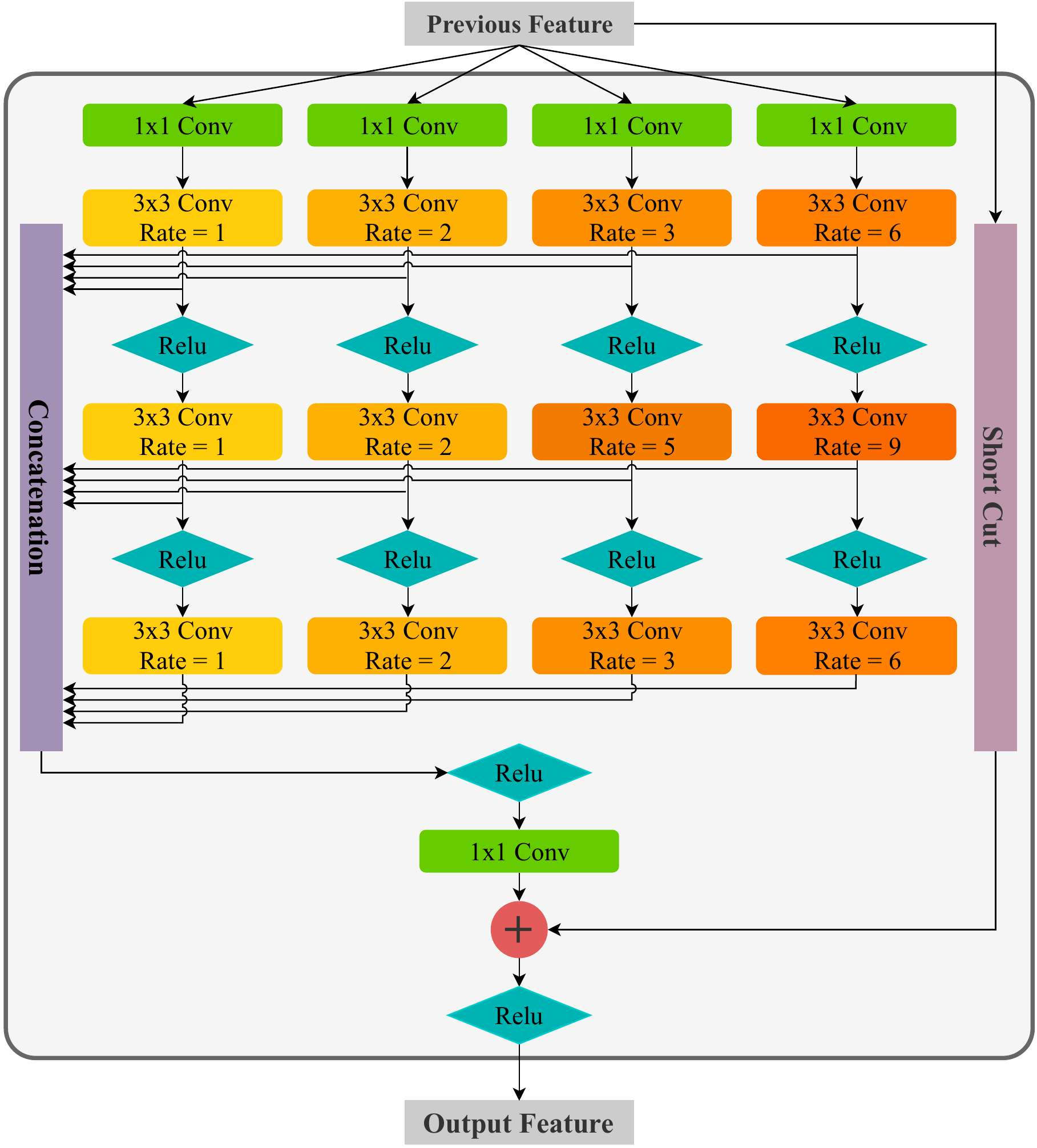}
\caption{The overall architecture of the {\ourmethod} block, where all $3 \times 3$ convolutions correspond to dilated convolution operations with adjustable dilated rates.}
  \label{fig:block}
\end{figure}

In this section, we first introduce our {\ourmethod} block, explore the details of its structure, and demonstrate the advantages of our design. Afterwards, we instantiate our {\ourmethod} block by showing how to integrate it in SSD~\cite{liu2016ssd} as {\ournet}. Without lose of generality, we describe how to apply our {\ourmethod} to other state-of-the-art object detectors.

\subsection{How {\ourmethod} is Constituted}
\label{sec:block-overview}
\input{method-block.tex}

\input{exp-bigtable.tex}

\subsection{When {\ourmethod} Meets SSD}
\label{sec:ssd-arch}
\input{method-architecture.tex}

\subsection{Applying {\ourmethod} to Other Detectors}
\label{sec:general-arch}
\input{method-general.tex}

%% file: method-block.tex
Our goal is to design a fast yet powerful feature fusion module to facilitate multi-scale object detection. We inherit the merits of feature fusions to enable the {\ourmethod} block to have finer granularity of fused representation. Meanwhile, to remedy the well-known issues of feature fusion (e.g., inefficiency with large kernel sizes), we adopt the dilated convolutions in {\ourmethod} to reduce the complexity of block, thus achieving the goal of fast computation. As depicted in Fig.~\ref{fig:block}, the {\ourmethod} block is a \textbf{multi-level multi-branch dilated convolution} block which contains two major attributes: multi-level multi-branch structure for comprehensive and fine-grained feature fusion and dilated convolutions for efficient inference. In what follows, we illustrate these designs in detail.

\input{method-block-structure}

\begin{algorithm}[!t]
    \centering
	\caption{Process of {\ourmethod} block}
	\label{alg:block_op}
	\begin{algorithmic}[1]
		\small
		\Require $\mathbf{X}_{\mathrm{pre}}$ with shape $(b, h, w, c_{\mathrm{pre}})$
		\Ensure $\mathbf{X}_{\mathrm{out}}$ with shape $(b, h, w, c_{\mathrm{out}})$
		\For{$l=1, \dots, 3$}
			\For{$r=1, \dots, 4$}
                \If{$l > 1$}
    				\State $\mathbf{X}_{l,r} =$ \texttt{ReLu}$(\mathbf{X}_{l-1,r})$	
                \Else
            		\State $\mathbf{X}_{1,r} =$  \texttt{Conv1\_1}$(\mathbf{X}_{pre})$
                \EndIf
                \State $\mathbf{X}_{l,r} =$ \texttt{Conv3\_3}$(\mathbf{X}_{l, r})$
				\Comment{$\mathbf{X}_{l,r}$ has $c_{\mathrm{pre}} / 4$ channels}
			\EndFor
		\EndFor
		\State $\mathbf{X}_{\mathrm{concate}} =$ \texttt{Concate}$(\{\mathbf{X}_{l,r}\}_{1 \leq l \leq L, 1 \leq r \leq R})$
        \State		
        \Comment{$\mathbf{X}_{\mathrm{concate}}$ has $3c_{\mathrm{pre}}$ channels}
		\State $\mathbf{X}_{\mathrm{shortcut}} =$ \texttt{ShortCut}$(\mathbf{X}_{\mathrm{pre}})$
        \State
		\Comment{$\mathbf{X}_{\mathrm{shortcut}}$ has $c_{\mathrm{out}}$ channels}
		\State $\mathbf{X}_{\mathrm{out}} =$ \texttt{ReLu}$($\texttt{Conv1\_1}$($\texttt{ReLu}$(\mathbf{X}_{\mathrm{concate}}))$ $+ \mathbf{X}_{\mathrm{shortcut}})$
	\end{algorithmic}
\end{algorithm}

\begin{figure*}[t!]
  \centering
  \includegraphics[width=0.8\linewidth]{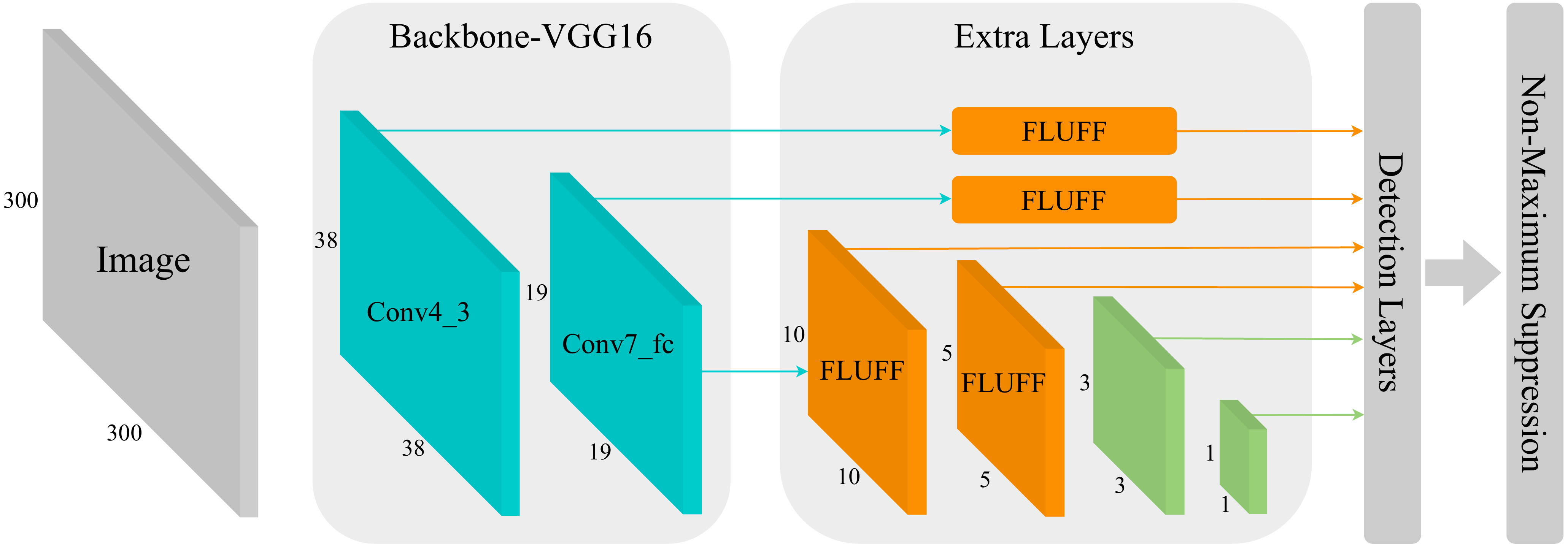}
  \caption{Architecture of {\ournet}300.}
  \label{fig:arch}
\end{figure*}

\input{method-block-dconv}

\input{method-block-alg}

%% file: method-block-structure.tex
\subsubsection{Multi-level multi-branch structure}
{\ourmethod} is composed of multiple parallel convolution branches with a short cut, in which each branch contains multiple dilated convolution layers with kernel size as $3 \times 3$ and followed by ReLU~\cite{nair2010rectified} activations. As clearly demonstrated in related works~\cite{szegedy2015going,liu2018receptive,li2019scale}, 
multi-branch structure can explicitly enlarge the scale variations of receptive fields by adapting varying sized convolutions in parallel branches and effectively generate multi-scale features with finer granularity, which are fused later by branch-wise concatenation and fed to the downstream object detection layers.

Meanwhile, we should consider expanding both the depth and the width of the network instead of width solely to boost the representational ability of the block. Generally, smaller receptive fields in shallower layers are more desirable for detecting objects with simpler appearances, and vice versa~\cite{zhao2019m2det}. Considering instances of similar scales with different appearances, existing works may fail to detect them since they usually exploit the same level of feature maps (similar instance sizes) and ignore the appearance differences. Thus, we equip {\ourmethod} with multi-level design and fuse features by level-wise concatenation afterward.

Unlike previous approaches, feature fusions of {\ourmethod} are laid out in cascade and parallel simultaneously (e.g., RFBNet applies no level-wise fusion~\cite{liu2018receptive}). {\ourmethod} not only benefits from multi-branch fusion but uses the multi-level fusion to further reinforce the appearance representation to achieve the comprehensive and extremely fine-grained feature fusion with the state-of-the-art performance.

Moreover, most of the existing feature fusion modules function on multiple crucial feature maps to simulate \textit{body-level} feature fusion as a \textit{coarse-grained} fusion method (e.g., TridentNet conducts fusions on the backbone and cannot port to other networks~\cite{li2019scale}). On the contrary, we apply the {\ourmethod} block on specific sizes of feature maps independently as a \textit{head-level} fusion with finer granularity. Compared with the existing modules, {\ourmethod} greatly intensifies the discriminative power of multi-scale features maps for encoding the multi-scale instances and hence accomplishing the goal of fine-grained feature fusion.

%% file: method-block-dconv.tex
\subsubsection{Dilated convolutions}
Traditional multi-scale feature fusion is achieved by convolutions with different kernel sizes. Inevitably, convolutions with larger kernel sizes tend to have more parameters, and the range of receptive fields extracted from this approach is often limited.
Recently, dilated convolutions~\cite{holschneider1990real,yu2015multi} have been shown to effectively magnify the kernel sizes of convolutions and increase the receptive fields without extra cost, and they have been widely adopted in tons of works like semantic segmentation to capture larger context information and speeding up the inference at the same time.
Furthermore, dilated convolutions with different dilated rates can efficiently resample the features at different scales and rapidly generate multi-scale receptive fields from the feature maps of fixed size~\cite{chen2017deeplab,liu2018receptive,li2019scale}.
Therefore, we employ dilated convolutions in {\ourmethod} to strengthen the multi-scale feature representation by setting various dilated rates in both branch-wise and level-wise and reduce the complexity of {\ourmethod}. Incorporating with multi-level multi-branch structure, {\ourmethod} utilizes the dilated convolutions to further raise the power of multi-scale feature extraction and elevate the efficiency of fine-grained feature fusion. Specifically, as shown in Fig.~\ref{fig:block}, dilated rates of convolutions have been adaptively adjusted on both horizontal and vertical aspects.

%% file: method-block-alg.tex
\subsubsection{Process of the {\ourmethod} block}
{\ourmethod} takes the input features $\mathbf{X}_{\mathrm{pre}}$ and returns the output features $\mathbf{X}_{\mathrm{out}}$.
To elaborate the details of the whole computation process of {\ourmethod} as described in Algorithm~\ref{alg:block_op}, we establish the notations as follows:
\begin{itemize}
	\item Let $\mathbf{X}$ denote the features with the shape of $(b, h, w, c)$, where $b, h, w, c$ refer to batch size, height, width and the number of channels, respectively.
	\item Let $\{\mathbf{X}_{l,r}\}_{1 \leq l \leq L, 1 \leq r \leq R}$ denote the processed features at level $l$ in branch $r$, where each {\ourmethod} block has $L$ levels and $R$ branches. The example of {\ourmethod} in Fig.~\ref{fig:block} has $L=3$ levels and $R=4$ branches.
	\item Let \texttt{ReLu} denote the rectified linear unit~\cite{nair2010rectified} activation, \texttt{Concate} refer to concatenation operation, and \texttt{ShortCut} correspond to shortcut process.
	\item Let \texttt{Conv1\_1} and \texttt{Conv3\_3} refer to $1 \times 1$ convolutions and $3 \times 3$ dilated convolutions, respectively.
\end{itemize}

%% file: exp-bigtable.tex
\begin{table*}[t!]
  \centering
    \caption{Comparisons on the MS COCO \textit{test-dev} set. {\ournet}, SSD, RFBNet, EfficientDet-D0, and M2Det were measured on Nvidia Titan X (Pascal), while other results stemmed from the references. The real-time (FPS $\geq$ 10) models reported in the table are \textbf{sorted ascendingly by FPS}.}
	\resizebox{0.9\linewidth}{!}
	{
    \begin{tabular}{llccccccc}
    \toprule
    \multirow{2}*{Model}&\multirow{2}*{Backbone}&
    \multirow{2}*{FPS}&\multicolumn{3}{c}{Avg. Precision, IoU}&\multicolumn{3}{c}{Avg. Precision, Area}\cr
    \cmidrule(lr){4-6} \cmidrule(lr){7-9}
    &&&0.5:0.95&0.5&0.75& S & M & L\\
    \midrule
    Faster-by-G-RMI~\cite{Huang_2017_CVPR}&Inception-ResNet-v2
    &--&34.7&55.5&36.7&13.5&38.1&52.0\\
    Faster+++~\cite{He_2016_CVPR}&ResNet101&0.29&34.9&55.7&37.4&15.6&38.7&50.9\\
    Mask-R-CNN~\cite{He_2017_ICCV}&ResNext101-FPN&4.7&37.1&60.0&39.4&16.9&39.9&\textbf{53.5}\\
    Faster-w-FPN~\cite{Lin_2017_CVPR}&ResNet101-FPN&4.1&36.2&59.1&39.0&18.2&39.0&48.2\\
    Faster~\cite{NIPS2015_5638}&VGG&6.8&24.2&45.3&23.5&7.7&26.4&37.1\\
    R-FCN-Deformable~\cite{Dai_2017_ICCV}&ResNet101&8&34.5&55.0&--&14.0&37.7&50.3\\
    R-FCN~\cite{NIPS2016_6465}&ResNet101&9&29.9&51.9&--&10.8&32.8&45.0\\
    \cmidrule(lr){1-9}
    RetinaNet800~\cite{Lin_2017_ICCV}&ResNet101-FPN& 5&39.1&59.1&42.3&21.8& 42.7&50.2\\
    DSSD513~\cite{2017arXiv170106659F}&ResNet101&5&33.2&53.3&35.2&13.0&35.4&51.1\\
    \cmidrule(lr){1-9}
    RetinaNet500~\cite{Lin_2017_ICCV}&ResNet101-FPN&11&34.4& 53.1& 36.8& 14.7& 38.5& 49.1\\
    M2Det512~\cite{zhao2019m2det}&VGG&13&37.6&56.6&40.5&18.4&43.4&51.2 \\
    EfficientDet-D0 (512)~\cite{tan2020efficientdet}&EfficientNet-B0&16&33.8&52.2&35.8&-&-&- \\
    \textbf{{\ournet}800 (Ours)}&VGG& \textbf{18}& \textbf{38.3}& 58.0& 41.6& 21.8& 43.6& 50.1\\
    \cmidrule(lr){1-9}
    YOLOv3 (608)~\cite{2018arXiv180402767R}&Darknet53&20&33.0&57.9&34.4&18.3&35.4&41.9 \\
    M2Det320~\cite{zhao2019m2det}&VGG&21&33.5&52.4&35.6&14.4&37.6&47.6 \\
    RefineDet512~\cite{Zhang_2018_CVPR}&VGG&22& 33.0&54.5&35.5&16.3&36.3&44.3 \\
    RFBNet512-E~\cite{liu2018receptive}&VGG&29& 34.4& 55.7& 36.4& 17.2& 37.0& 47.6\\
    RFBNet512~\cite{liu2018receptive}&VGG&32& 33.8& 54.2& 35.9& 16.2& 37.1& 47.4\\
    \textbf{{\ournet}512 (Ours)}&VGG& \textbf{32}& \textbf{35.8}& 56.3& 38.2& 18.0& 39.2& 50.1\\
    \cmidrule(lr){1-9}
    SSD512~\cite{liu2016ssd}&VGG&36& 28.8& 48.5& 30.3& 10.9&31.8& 43.5\\
    RefineDet320~\cite{Zhang_2018_CVPR}&VGG&39&29.4&49.2&31.3&10.0&32.0&44.4 \\
    YOLOv2 (544)~\cite{Redmon_2017_CVPR}&Darknet19&40&21.6&44.0&19.2&5.0&22.4&35.5\\
    RFBNet300~\cite{liu2018receptive}&VGG&50& 30.3& 49.3& 31.8& 11.8& 31.9& 45.9 \\
    \textbf{{\ournet}300 (Ours)}&VGG& \textbf{50}& \textbf{31.6}& 50.7& 33.3& 12.5& 33.5& 47.8\\
    SSD300~\cite{liu2016ssd}&VGG&83&25.1&43.1&25.8&6.6&25.9&41.4\\
    \bottomrule
    \end{tabular}}
  \label{tab:coco_performance_comparison}
\end{table*}

%% file: method-architecture.tex
Motivated by the success of recent works~\cite{liu2018receptive,zhao2019m2det} which conducted effective upgrades on a powerful multi-scale one-stage object detector -- SSD, we integrated our {\ourmethod} into it to build a fast yet accurate object detector named \textbf{{\ournet}} to verify the validity of {\ourmethod}, where Fig.~\ref{fig:arch} depicts the overall architecture of {\ournet}. Leveraging our {\ourmethod} block is remarkably simple, and due to the great portability of {\ourmethod}, we have retained most of the structures of the SSD with several minor modifications.

\paragraph{Base network} To ensure the relatively fair comparisons with other methods, we adopt the same pre-trained lightweight backbone VGG16~\cite{simonyan2014very} as used in SSD. Following RFBNet~\cite{liu2018receptive}, VGG16's fc6 and fc7 are converted to convolutions with sub-sampling parameters, pool5 is changed to \mbox{3x3-s1}, all dropout layers and fc8 are eliminated, and dilated convolutions are embedded to fill the ``holes''.

\paragraph{Multi-scale feature maps} SSD is a typical example of the object detector with multi-scale feature maps. Originally, it extends the truncated backbone network by adding additional convolutional layers in a cascade fashion, where the sizes of these layers are progressively decreased, which enables SSD to detect objects at various scales. 
Particularly, these multi-scale feature maps mainly fall into two categories: ones extracted from the backbone (e.g., conv4\_3, conv7\_fc) and ones produced by extra feature layers.
To maximize the representational ability of the {\ourmethod} block, we applied it to these two categories of multi-scale features in a slightly different way. 
Primarily, we utilized the {\ourmethod} block to post-process the features generated by the backbone, and traditional convolutions in extra feature layers are substituted by {\ourmethod}. In this case, our {\ourmethod} blocks could act on feature maps of different sizes/levels separately to conduct the efficient and effective multi-scale feature fusion on \textit{head-level}, and achieve the goal of fine-grained feature fusion.
The overall architecture is illustrated in Fig.~\ref{fig:arch}.

%% file: method-general.tex
Our {\ourmethod} block can be deployed into other detectors (e.g., YOLO~\cite{Redmon_2016_CVPR}) as an essential and powerful upgrade to alleviate the drawbacks of coarse-grained feature fusion of these networks, thus further enhancing the ability of feature representation and improving the performance. Only two simple steps need to be performed to employ our {\ourmethod}:
1) Use the {\ourmethod} block to post-process the features extracted by the backbone, and 2) replace traditional convolutions with {\ourmethod} for features produced by extra layers.

%% file: exp.tex
We conducted extensive experiments on MS COCO~\cite{lin2014microsoft} and PASCAL VOC~\cite{everingham2007pascal}, which have 80 and 20 categories of various-scale objects, respectively. The metric for assessing the detection accuracy is the mean average precision (mAP) given several specific intersection over union (IoU) thresholds on MS COCO. 
Meanwhile, we calculated mAP with IoU of $0.5$ solely on PASCAL VOC.
To measure the efficiency of the detectors, we defined $\tinfer$ of models as ${\tinfer}={\tpred}+{\tnms}$,
where $\tinfer$ refers to the inference time and represents the total time needed to generate final outputs, $\tpred$ denotes the prediction time which indicates the actual running time and intuitively reflects the complexity of the models, and $\tnms$ corresponds to the time consumed by the non-maximum suppression (NMS). Furthermore, frame per second ($\mathrm{FPS}=1/{\tinfer}$) is reported in our experiments as well.

\subsection{MS COCO}
\input{exp-coco.tex}

\label{sec:exp-coco}

\begin{figure*}[t!]
  \centering
  \includegraphics[width=0.7\linewidth]{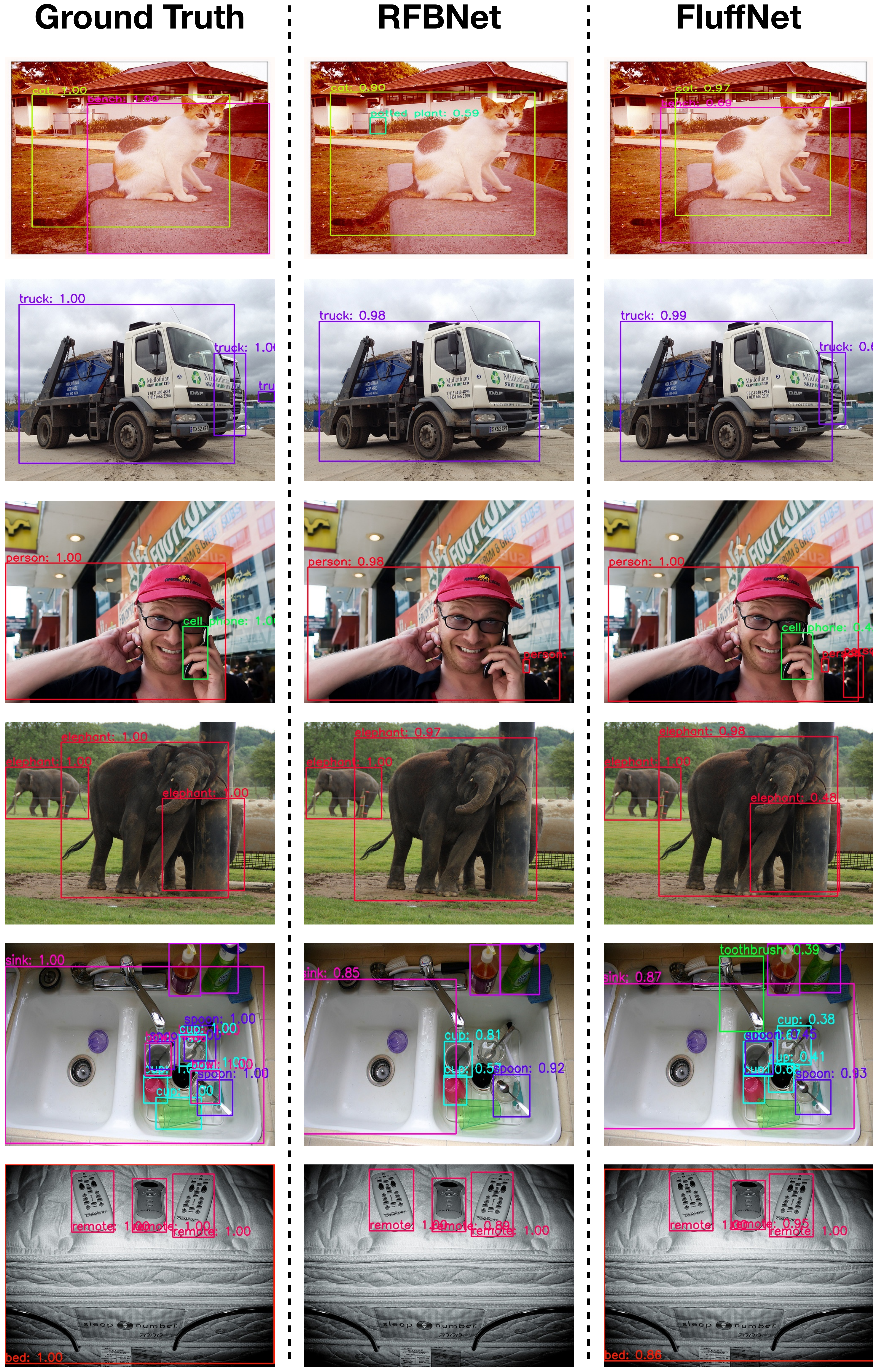}
\caption{Qualitative comparisons between FluffNet and RFBNet, the columns are ground truth, results of RFBNet and FluffNet on MS COCO, respectively.}
  \label{fig:vis_mscoco}
\end{figure*}

\subsection{Pascal VOC}
\input{exp-voc.tex}
\label{sec:exp-voc}

\subsection{Applying {\ourmethod} to Other Detectors}
\input{fluff_detectors.tex}

\begin{table}[t!]
\centering
\caption{Comparisons of various multi-scale designs on the PASCAL VOC \textit{test} set with the input size of $300 \times 300$.}
\resizebox{0.99\linewidth}{!}
{
\begin{tabular}{l|c|cccc|c}
\toprule
Components&SSD&ANet&BNet&CNet&DNet&{\ournet}\\
\midrule
Multi-level&~~&\ding{51}&~~&\ding{51}&\ding{51}&\ding{51}\\
Multi-branch&~~&~~&\ding{51}&\ding{51}&\ding{51}&\ding{51}\\
Dilated-conv&~~&~~&~~&~~&\ding{51}&\ding{51}\\
ReLU&~~&~~&~~&~~&~~&\ding{51}\\
\cmidrule(lr){1-7}
\textit{mAP(\%)}&77.2&78.7&78.8&79.4&79.7&\textbf{80.8}\\
${\tpred}$&5ms&26ms&28ms&36ms&11ms&\textbf{11ms}\\
\bottomrule
\end{tabular}
}
\label{tab:ablation_study}
\end{table}

\begin{table}[t!]
\centering
\caption{Comparisons of various fusions on the PASCAL VOC \textit{test} set with the input size of $300 \times 300$. 
All networks have multi-level multi-branch structures without ReLU activations between adjacent convolutional layers.
}
\resizebox{0.8\linewidth}{!}
{
\begin{tabular}{l|cccc}
\toprule
Components&CNet& & & DNet \\
\midrule
Max-pooling&~~&\ding{51}&~~&~~\\
Avg-pooling&~~&~~&\ding{51}&~~\\
Dilated-conv&~~&~~&~~&\ding{51}\\
\cmidrule(lr){1-5}
\textit{mAP(\%)}&79.4&79.2&79.4&\textbf{79.7}\\
${\tpred}$&36ms&10ms&11ms&11ms\\
\bottomrule
\end{tabular}
}
\label{tab:ablation_study_two}
\end{table}

\subsection{Ablation Study}
\input{exp-ablation.tex}

\label{sec:exp-abl}

%% file: exp-coco.tex
\paragraph{Experimental settings}
We trained all models on the \textit{trainval35k} set (\textit{train} + \textit{val35k}) and tested them on \textit{test-dev}. We set batch size to 32, learning rate to $4\times10^{-3}$, and weight decay to $5\times10^{-4}$. We adopted the ``warm-up'' strategy at the beginning of training: in the first 5 epochs, the learning rate gradually increased from $10^{-6}$ to $4\times10^{-3}$, and then decreased by 10 times in the 90th, 120th, and 140th epoch respectively.
Regarding the dilated rates of {\ourmethod}, they were initialized as $\mathrm{init_{rate}}=[1, 2, 3, 6]$ at all levels. The {\ourmethod} blocks utilized to post-process the backbone features kept the $\mathrm{init_{rate}}$, while the {\ourmethod} blocks in extra feature layers adaptively revised dilated rates according to the input size, i.e., {\ourmethod} adjusted the dilated rates at $l$th level to the rounded integers of $(1+0.25l)\times\mathrm{init_{rate}}$, where $l=1, 2$ for {\ournet}300, and $l=1, 2, 3$ for {\ournet}512. In {\ourmethod}, all the dilated rates were fixed and the total levels were 3 during the training. {\ournet} was converged around 280 epochs.

\paragraph{Comparisons with state-of-the-art detectors}
As illustrated in Table~\ref{tab:coco_performance_comparison}, our {\ourmethod} has achieved state-of-the-art performance on MS COCO, i.e., it obtained the fastest inference speed while outperforming other competitive detectors in terms of mAP. Admittedly, RetinaNet800 accomplished the highest mAP, it was too slow to cross the chasm to overtake other real-time (i.e., FPS $\geq$ 10) detectors as a practical application, not to mention our {\ournet}. SSD300 obtained the best efficiency regarding the inference speed; however, it is obstructed by an unbridgeable gap with {\ournet}300 in terms of mAP.
Beyond that, we roughly grouped real-time detectors into three subsets, in which each contained {\ournet} of specific input size (e.g., 800, 512 and 300). Concretely, {\ournet}800 is much faster than RetinaNet500 and M2Det512 and acquired the best accuracy; {\ournet}512 achieved highest speed of 32 FPS with greatest mAP of 35.8\% compared with M2Det320, RFBNet512, etc., and {\ournet}300 is at the top speed of 50 FPS while maintaining the supreme mAP of 31.6\% among SSD512, RefineDet320, and so on. More importantly, {\ournet} gained a lot through the {\ourmethod} block since mAPs of small and large objects were much higher than other real-time detectors, which explicitly evidenced the validity and feasibility of our designs for fine-grained multi-scale feature fusion.

\begin{figure}[ht!]
  \centering
  \includegraphics[width=0.9\linewidth]{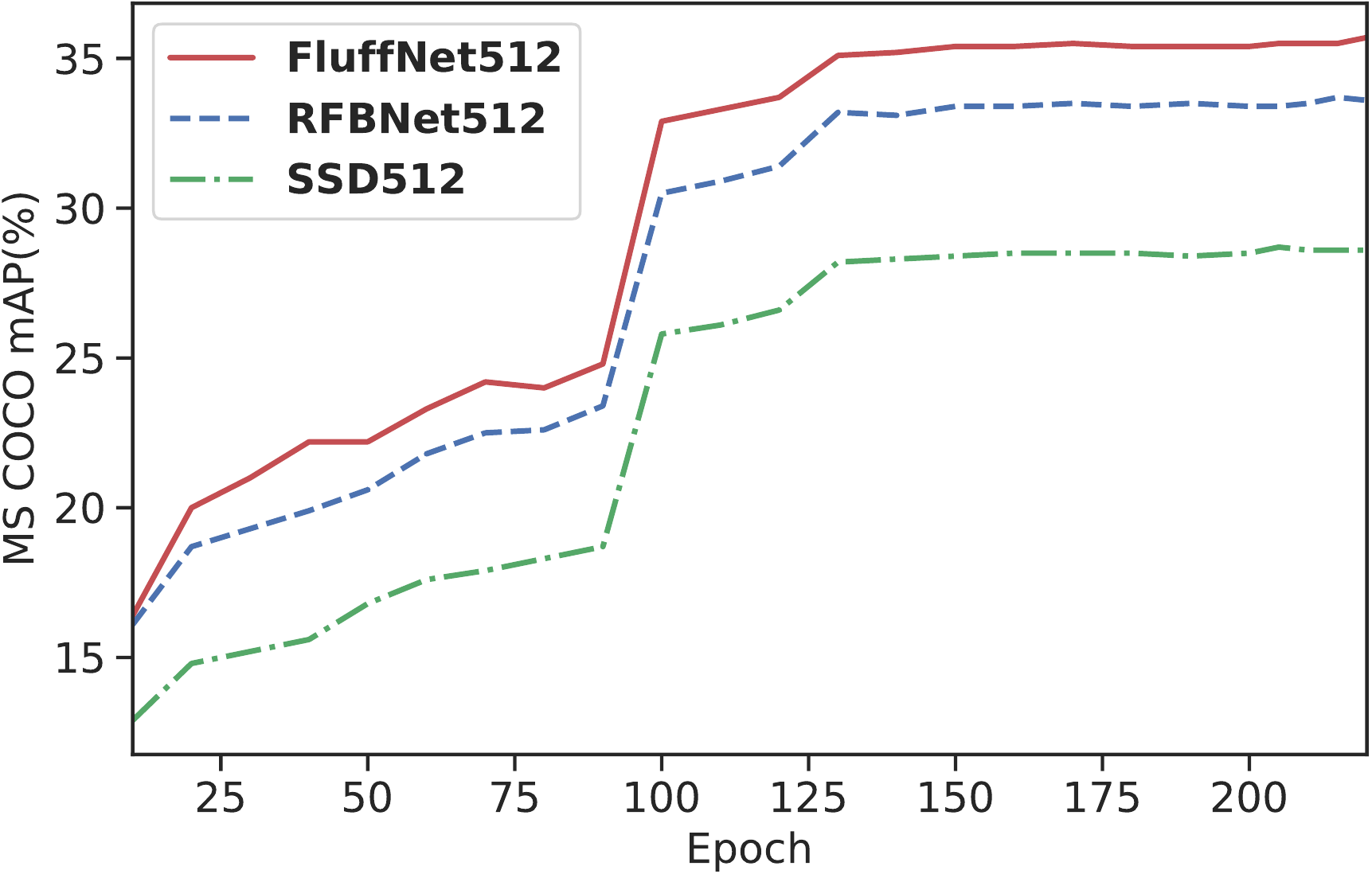}
\caption{Epoch vs. accuracy (mAP) curves on MS COCO \textit{minival} set for SSD512, RFBNet512 and FluffNet512}
  \label{fig:ecpoch_map_ssd}
\end{figure}

\paragraph{Comparisons with SSD-based detectors}
Similarly, existing works such as RFBNet and M2Det were built on top of the SSD, as a result, we carried out more detailed comparative trials with these models. As shown in Table~\ref{tab:comparison_with_other_detector}, comparisons were carried out among {\ournet}, RFBNet and M2Det. Generally speaking, with the increase of input size, models become larger. When {\ournet} enlarged input size from 300 to 800, the number of its parameters slightly increased from 58M to 63M. Moreover, with the same inference speed, {\ournet}300 and {\ournet}512 obtained higher mAP compared with RFBNet300 and RFBNet512 respectively, and {\ournet}800 achieved best accuracy among M2Det320 and M2Det512 and kept the higher efficiency.
In addition, Fig.~\ref{fig:ecpoch_map_ssd} shows the epoch-accuracy curves of SSD512, RFBNet512 and FluffNet512 on MS COCO \textit{minival} set. It is undeniable that FluffNet512 consistently excelled other two detectors, which demonstrated that {\ourmethod} block could strengthen the performance of networks for multi-scale object detection.

\begin{table}[t!]
\centering
\caption{Comparisons of SSD-based models on the MS COCO \textit{test-dev}.}
\resizebox{0.95\linewidth}{!}
{
\begin{tabular}{llcccc}
\toprule
Detectors&\#params&${\tpred}$&${\tnms}$&${\tinfer}$&mAP(\%)\\
\midrule
RFBNet300&42M&7ms&13ms&20ms&30.3\\
\textbf{{\ournet}300}&58M&11ms&9ms&20ms&31.6\\
\cmidrule(rl){1-6}
RFBNet512&44M&9ms&22ms&31ms&33.8\\
\textbf{{\ournet}512}&60M&15ms&16ms&31ms&35.8\\
\cmidrule(rl){1-6}
M2Det320&126M&45ms&3ms&48ms&33.5\\
\textbf{{\ournet}800}&63M&19ms&37ms&56ms&38.3\\
M2Det512&126M&65ms&13ms&78ms&37.6\\
\bottomrule
\end{tabular}}
\label{tab:comparison_with_other_detector}
\end{table}

\paragraph{Qualitative comparisons with RFBNet}
We further highlighted the advantages of {\ourmethod} block by showing the predictions of RFBNet with {\ournet} since RFBNet was the most related approach in terms of efficiency and accuracy. As depicted in Fig.~\ref{fig:vis_mscoco}, the visual variances between the predictions of RFBNet and {\ournet} sufficiently demonstrated that the {\ournet} outperformed RFBNet, especially on huge (e.g., top row) and tiny (e.g., second row) cases.

%% file: exp-voc.tex
\paragraph{Experimental settings}
We trained models on a combination of VOC2007 \textit{trainval} set and VOC2012 \textit{trainval} set and tested them on the VOC2007 \textit{test} set.
All models were trained in a batch size of 32, and the learning rate was initialized at $4\times10^{-4}$. The total number of training epochs in our experiments was set to 300, and weight decay and momentum were set as 0.0005 and 0.9 respectively. The {\ournet} was converged around 235 epochs. Moreover, the dilated rates and total levels of {\ourmethod} blocks were applied the same procedure as mentioned in MS COCO to guarantee the great ability of multi-scale feature fusion.

\begin{table}[t!]
\centering
\caption{Comparisons on the PASCAL VOC \textit{test} set. The results of {\ournet}, SSD, RFBNet, and M2Det were measured on Nvidia Titan X (Pascal), and other results stemmed from the references. Real-time (FPS $\geq$ 10) methods are \textbf{sorted ascendingly by FPS}.}

\resizebox{0.9\linewidth}{!}
{
\begin{tabular}{llcc}
\toprule
Method&Backbone&FPS&mAP(\%)\\
\midrule
Faster+++&ResNet101&5&76.4\\
Faster&VGG&7&73.2\\
R-FCN-Deformable&ResNet101&9&80.5\\
\cmidrule(lr){1-4}
DSSD513&ResNet101&6&81.5\\
DSSD321&ResNet101&10&78.6\\
RefineDet512&VGG&24&81.8\\
YOLOv2 (544)&Darknet19&40&78.6\\
RefineDet320&VGG&40&80.0\\
\textbf{{\ournet}512 (Ours)}&VGG&\textbf{56}&\textbf{82.5}\\
\cmidrule(lr){1-4}
RFBNet512&VGG&67&82.2\\
\textbf{{\ournet}300 (Ours)}&VGG&\textbf{71}&\textbf{80.8}\\
SSD512&VGG&83&79.8\\
RFBNet300&VGG&95&80.4\\
SSD300&VGG&125&77\\
\bottomrule
\end{tabular}}
\label{tab:voc_performance_comparison}
\end{table}

\paragraph{Comparisons with state-of-the-art detectors}
Table~\ref{tab:voc_performance_comparison} presents the comparisons among {\ournet} and other detection methods. Similarly, we divided all real-time methods into two sub-groups according to FPS in which we could clearly notice that {\ournet} achieved the state-of-the-art performance by exceeding existing detectors with a remarkable gap. In particular, {\ournet} obtained fastest speed of 56 FPS and highest detection accuracy of 82.5\% compared with DSSD513, DSSD321, YOLOv2, etc. Moreover, {\ournet} reached the peak of the mAP of 80.4\% while keeping the comparable computation efficiency compared with competitive baselines such as RFBNet512, RFBNet300.

\paragraph{Performance variations on two datasets}
Apparently, performance were dissimilar between two benchmarks due to the distinctions of scale distributions. Specifically, the ratios of small, medium and large instances in MS COCO and PASCAL VOC are 41.5\%/34.3\%/24.2\% and 4.7\%/29.0\%/66.3\%, respectively, and this variation of size distributions led to a different performance on two benchmarks since {\ourmethod} is more adept at small object detection.

%% file: fluff_detectors.tex
\begin{table*}[t!]
  \centering
  \caption{Comparisons of YOLOv3 and Faster-RCNN with or without {\ourmethod} block on MS COCO \textit{minval} set.}
  \resizebox{0.75\linewidth}{!}
	{
    \begin{tabular}{llccccccc}
    \toprule
    \multirow{2}*{Model}&\multirow{2}*{Backbone}&
    \multirow{2}*{FPS}&\multicolumn{3}{c}{Avg. Precision, IoU}&\multicolumn{3}{c}{Avg. Precision, Area}\cr
    \cmidrule(lr){4-6} \cmidrule(lr){7-9}
    &&&0.5:0.95&0.5&0.75& S & M & L\\
    \midrule
    \textbf{YOLOv3-Fluff (608)}&Darknet53&40&40.7&60.3&44.0&22.6&45.0&53.5\\
    YOLOv3 (608)&Darknet53&45&39.1&59.1&42.4&22.3&44.0&50.3\\
    \midrule
    \textbf{Faster-RCNN-Fluff (600)}&ResNet101&7&32.1&52.7&33.8&12.5&36.4&49.0\\
    Faster-RCNN (600)&ResNet101&8&30.6&51.2&31.9&11.6&34.9&47.3\\
    \bottomrule
    \end{tabular}}
  \label{tab:yolo_faster_rcnn_comparison}
\end{table*}

\begin{figure}[h]
  \centering
  \includegraphics[width=0.9\linewidth]{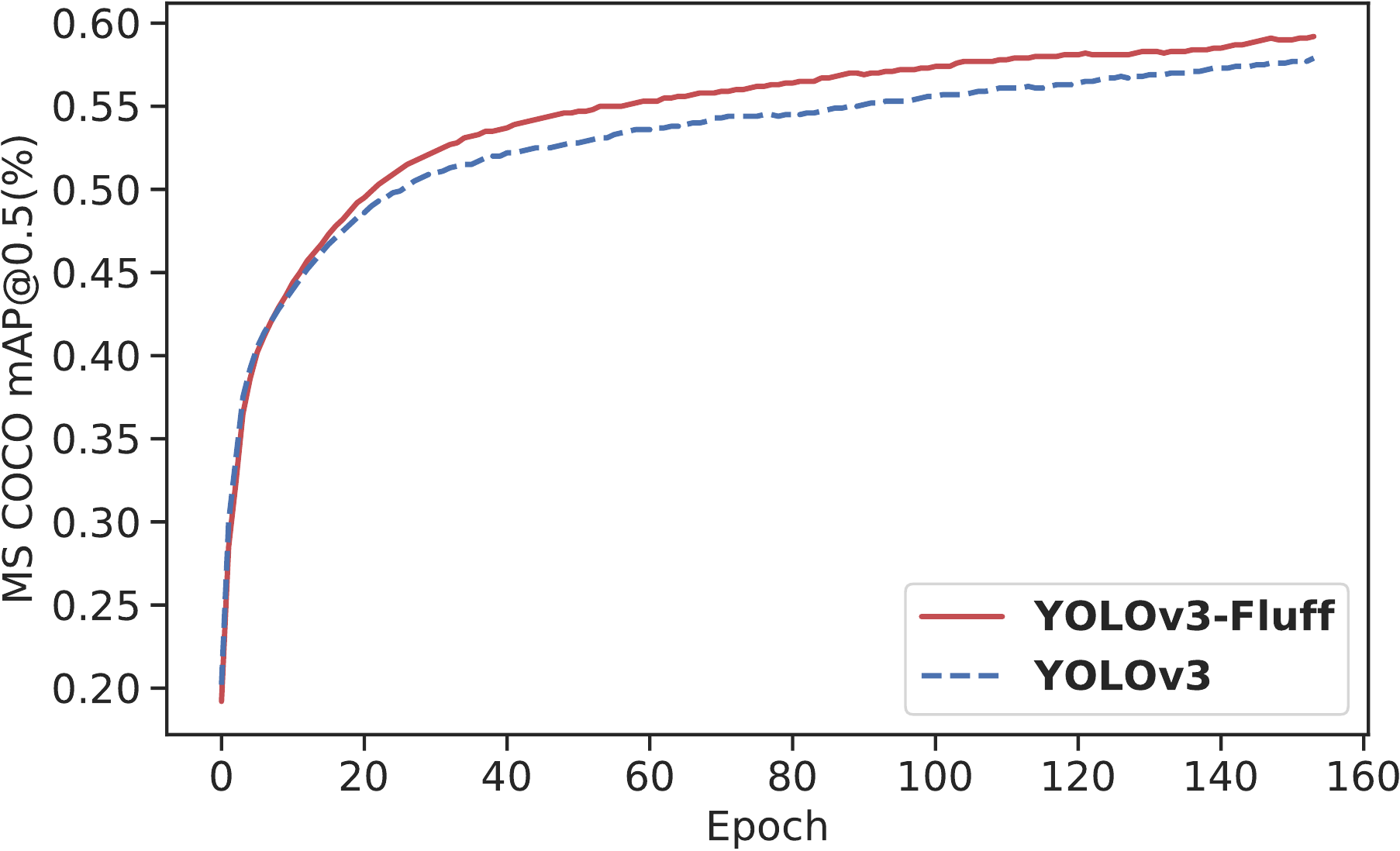}
\caption{Epoch vs. accuracy (mAP) curves on MS COCO \textit{minival} set for YOLOv3 and YOLOv3-Fluff.}
  \label{fig:ecpoch_map_yolov3}
\end{figure}

\begin{figure}[h]
  \centering
  \includegraphics[width=0.9\linewidth]{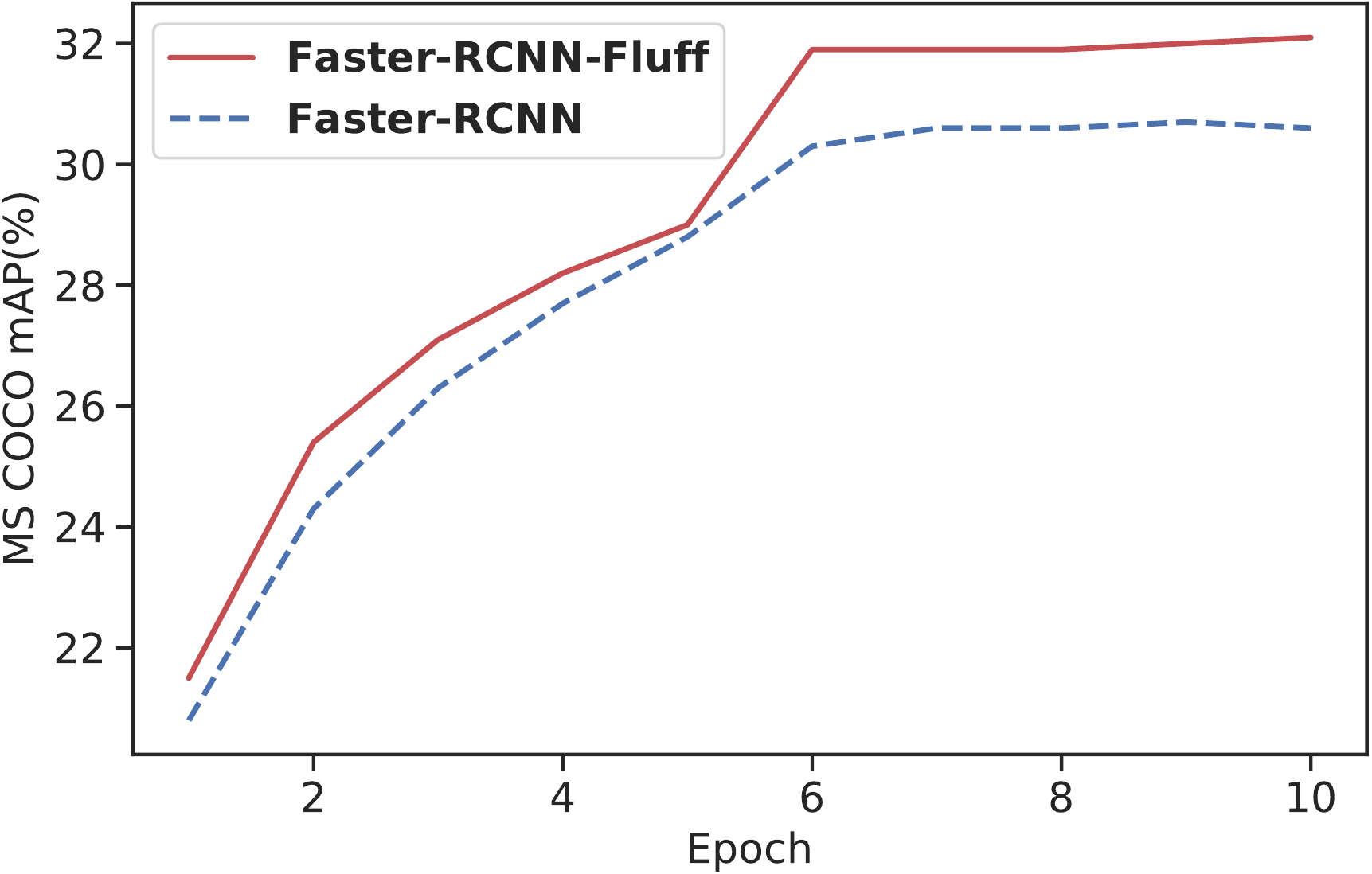}
\caption{Epoch vs. accuracy (mAP) curves on MS COCO \textit{minival} set for Faster-RCNN and Faster-RCNN-Fluff.}
  \label{fig:ecpoch_map_faster_rcnn}
\end{figure}

To demonstrate the generality and portability of {\ourmethod} block, we embedded {\ourmethod} block into a one-stage detector YOLOv3 as YOLOv3-Fluff and a two-stage detector Faster-RCNN~\cite{NIPS2015_5638} as Faster-RCNN-Fluff to enhance their performance. As depicted in Table~\ref{tab:yolo_faster_rcnn_comparison}, the comparisons between original and updated detectors on MS COCO \textit{minval} set have clearly evidenced the merits of {\ourmethod} block for the multi-scale feature fusion. Specifically, the accuracy (mAP) of YOLOv3 had been boosted by 1.6\% after embedding {\ourmethod} before the three prediction branches, the performance of Faster-RCNN had been improved by 1.5\% by integrating {\ourmethod} between the backbone and the detection layer. 
Moreover, Fig.~\ref{fig:ecpoch_map_yolov3} and Fig.~\ref{fig:ecpoch_map_faster_rcnn} illustrate the epoch-accuracy curves for YOLOv3, YOLOv3-Fluff, Faster-RCNN and Faster-RCNN-Fluff on MS COCO \textit{minival} set during the training process, respectively. Apparently, after applying {\ourmethod} to YOLOv3 and Faster-RCNN, the updated detectors consistently outperformed original ones from the beginning to the end of the training, which indicated the advances of {\ourmethod} block's design as well.

%% file: exp-ablation.tex
\paragraph{Multi-level and multi-branch}
We explored the effects of multi-level and multi-branch versus single-level and/or single-branch. The results are illustrated in Table~\ref{tab:ablation_study}.
Based on SSD, we built \textbf{BNet} with multi-branch single-level structure and traditional convolutions to achieve an mAP of 78.8\%.
Meanwhile, \textbf{ANet} with a multi-level single-branch block obtained an mAP of 78.7\%.
Finally, we combined multi-level multi-branch structure, which was similar to a small FPN block, and built \textbf{CNet} with an mAP of 79.4\%.
The performance of these networks evidenced the advantages of combining multi-level multi-branch structures.

\paragraph{Dilated convolutions}
To verify the effectiveness of dilated convolutions, we equipped max-pooling or avg-pooling to act as the dilated convolutions for the same size of the receptive field in \textbf{CNet} as different fusion methods.
In Table~\ref{tab:ablation_study_two}, all these modifications accelerated the original network.
Although the $\tpred$ of them were similar, the network with dilated convolution obtained the best mAP of 79.7\%.

\paragraph{ReLU activations}
To strengthen the nonlinearity of feature transformations of inter-level of {\ourmethod}, we employed the ReLU activations between consecutive layers, both after the concatenation and before the final output, to further boost the representational power of {\ourmethod}.
The comparisons between \textbf{DNet} and {\ourmethod} in Table~\ref{tab:ablation_study} strongly demonstrated the significance of ReLU, which enhanced the performance from 79.7\% to 80.4\% in terms of mAP.

%% file: conclusion.tex
In this paper, we propose the \textbf{F}ast object detection with \textbf{L}atticed m\textbf{U}lti-scale \textbf{F}eature \textbf{F}usion (\textbf{{\ourmethod}}) block -- a novel module with multi-level multi-branch fine-grained feature fusion designed to tackle the scale variation in multi-scale object detection. {\ourmethod} inherits the advantages of both multi-level and multi-branch fusions with dilated convolutions to achieve the goal of fast, comprehensive, and finer-grained feature fusions. We further embed {\ourmethod} into SSD to construct {\ournet}, which is a powerful real-time single-stage detector for multi-scale object detection. Empirically we present the state-of-the-art performance of {\ournet} on MS COCO and PASCAL VOC. Finally, we illustrate the great generality of {\ourmethod} by showing how to integrate it into other detectors in a surprisingly simple way as an essential module to facilitate the multi-scale object detection. 